\RequirePackage{fix-cm}
\pdfoutput=1
\documentclass[a4paper,10pt]{article}

\usepackage{authblk}
\usepackage{amsmath,amsfonts}
\usepackage{cleveref}
\usepackage{array}
\usepackage[caption=false,font=normalsize,labelfont=sf,textfont=sf]{subfig}
\usepackage{textcomp}
\usepackage{stfloats}
\usepackage{url}
\usepackage{verbatim}
\usepackage{graphicx}
\usepackage{cite}
\usepackage{algorithm}
\usepackage{algpseudocode}
\usepackage{float}
\usepackage{tikz}
\usepackage{forest}
\usetikzlibrary{fit}
\usepackage{multicol}

\providecommand{\keywords}[1]{\textbf{\textit{Keywords }} #1}

\begin{document}

\title{Combining Genetic Programming and Particle Swarm Optimization to Simplify Rugged Landscapes Exploration
}

\author[1]{Gloria Pietropolli}
\author[1]{Giuliamaria Menara}
\author[2]{Mauro Castelli}

\affil[1]{{\small Dipartimento di Matematica e Geoscienze, Università degli Studi di Trieste, H2bis Building, Via Alfonso Valerio 12/1, 34127 Trieste, Italy}
	
 	{\small \texttt{\{giuliamaria.menara, gloria.pietropolli\}@phd.units.it}}}

\affil[2]{{\small NOVA IMS, Universidade NOVA de Lisboa, Campus de Campolide, 1070-312, Lisboa, Portugal}

    {\small \texttt{mcastelli@novaims.unl.pt}}}

\maketitle

\begin{abstract}
Most real-world optimization problems are difficult to solve with traditional statistical techniques or with metaheuristics. The main difficulty is related to the existence of a considerable number of local optima, which may result in the premature convergence of the optimization process.
To address this problem, we propose a novel heuristic method for constructing a smooth surrogate model of the original function. The surrogate function is easier to optimize but maintains a fundamental property of the original rugged fitness landscape: the location of the global optimum.
To create such a surrogate model, we consider a linear genetic programming approach enhanced by a self-tuning fitness function.
The proposed algorithm, called the \emph{GP-FST-PSO Surrogate Model}, achieves satisfactory results in both the search for the global optimum and the production of a visual approximation of the original benchmark function (in the 2-dimensional case).
\end{abstract}


\section{Introduction}
An optimization problem consists in maximizing (or minimizing) a fitness function relative to a fixed search space.
This task can be particularly challenging in real-life applications, where one usually faces the presence of a considerable number of local optima, which might cause premature convergence of the method and a consequent failure to discover the global optimum.
Therefore, framing this matter in the context of a fitness landscape search, it is often necessary to explore a rugged multidimensional surface, which consistently turns the search for the global optimum into a challenging, time-consuming problem.
This irregularity prevents the adoption of simple optimization techniques such as gradient descent and hill-climbing, as they cannot successfully deal with such rugged landscapes. 

To mitigate the computational effort necessary for effectively exploring these challenging landscapes, we developed a \emph{surrogate model} (SM) approach. An SM consists of defining an approximated fitness function whose evaluation is typically less expensive than its original counterpart. SMs are often used to reduce the cost of expensive objective functions by acting as approximation models that mimic the behavior of the original function as closely as possible while being fast surrogates for time-consuming computer simulations. 
In a few words, an SM is an easily measurable mathematical model that approximates an expensive objective function as precisely as possible.

To build this SM-like method, our idea is to retain the appeal of \emph{genetic programming} (GP) \cite{koza1994genetic} algorithms, as they can handle challenging problems with high-quality designs, while enhancing computational efficiency.
GP is one of the most prominent evolutionary-based techniques, with the ability to evolve programs to solve specific problems, that belongs to the wider class of optimization techniques called \emph{population-based} methods \cite{luke11}.
GP employs a variable model representation that makes no assumptions about the classes of models that will be relevant to the problem at hand and places the fewest possible constraints on the space of models that are allowed to be explored throughout the evolutionary process.

Combining what we have discussed so far, the main purpose of this research is to find the global optimum of rugged landscapes by evolving, through a GP-based routine, a valid surrogate function to approximate our problem adequately. 
To achieve this objective, we implemented a novel method called the \emph{GP-FST-PSO Surrogate Model} that generates smooth surrogate approximations of the original landscapes, thus making their exploration (i.e., searching for the global optimum) straightforward.
The main idea presented in this work relies on a combination of a GP-based algorithm coupled with a self-tuning fitness function.
More specifically, to evaluate the fitness of the produced surrogate functions, we employ \emph{Fuzzy Self-Tuning Particle Swarm Optimization (FST-PSO)}~\cite{nobile2018fuzzy}, a setting-free version of particle swarm optimization (PSO). 
FST-PSO was selected because, being a self-tuning algorithm, it does not require any hyperparameters, and this facilitates the subsequent analysis of GP-FST-PSO.
To estimate GP-FST-PSO efficiency, we considered a set of benchmark functions that are commonly used to test the proficiency of various optimization methods, as they are characterized by high noise and ruggedness.
The proposed approach reveals its suitability for performing the proposed task. In particular, experimental results confirm its capability to find the global argminimum for all the considered benchmark problems and all the domain dimensions taken into account, thus providing an innovative and promising strategy for dealing with challenging optimization problems.

The paper is structured as follows: Section~\ref{sec: relwork} provides an overview of surrogate models and, in particular, of evolutionary-based techniques for implementing surrogate models.
Thereafter, Section~\ref{sec: matandmeth} reviews \emph{genetic programming} (Section~\ref{subsec: geneticprogramming}) and \emph{Fuzzy Self-Tuning Particle Swarm Optimization} (Section~\ref{subsec: fst-pso}). Section~\ref{subsec: gp-fst-pso} introduces the proposed technique for performing surrogate modeling by combining the two methods reviewed above. The benchmark functions and the experimental settings are described in Section~\ref{sec: expsett}, and the results of the experimental campaign are presented in Section~\ref{sec: results}.
Finally, Section~\ref{sec: conclusion} summarizes the main contributions of the paper and provides directions for further research.

\section{Related Works}
\label{sec: relwork}

The field of optimization has grown rapidly during the past few decades, and many new theoretical, algorithmic, and computational contributions have been proposed to solve various problems \cite{cavazzuti2013deterministic}, \cite{lin2012review}, \cite{gomes2019agent}, \cite{zhou2018comparative}.
Recent developments in the field of optimization methods can mainly be divided into \emph{deterministic} and \emph{heuristic} approaches.
Deterministic methods are based on a solid mathematical formulation and are commonly used to address simple optimization problems where the computational effort grows only polynomially with the dimensionality of the problem \cite{lin2012review}.
However, if the problem is NP-hard, the same effort grows exponentially, and even small problems can become unsolvable using these methods as they usually get trapped in local minima \cite{lin2012review}.
On the other hand, heuristic approaches are based on search strategies that incorporate some form of randomness that increases their robustness \cite{sorensen2018history}.
As a result, such algorithms are very effective in handling hard or ill-conditioned optimization problems \cite{papila2002shape}.
As in this study, we define a new heuristic SM-like approach by coupling genetic programming and particle swarm optimization, the remaining part of this section is organized as follows: Section~\ref{sec:GPandPSO} reviews existing works in which GP and PSO have been combined to address specific optimization tasks, and Section~\ref{sec:surrogate} summarizes important concepts and contributions concerning the use of GA for the definition of valuable surrogate models.

\subsection{Genetic Programming Coupled with Particle Swarm Optimization}
\label{sec:GPandPSO}
GP and PSO are two computational-based techniques, both taking inspiration from the progression of biological life.
GP is a global optimization algorithm that simulates the principles of the Darwinian theory of evolution, while PSO can mimic cooperation between individuals in the same group using swarm intelligence and the exchange of experiences from generation to generation.
Because of the peculiarities of GP and PSO, it is possible to effectively combine the global search characteristic of GP and the local search capability of PSO to avoid premature convergence and to improve the quality of solutions \cite{feng2006identification}.

Several works that have already exploited the integration of GP and PSO are worth mentioning: First, Poli et al. \cite{poli2005extending}, where the use of GP aims to extend PSO models by including biology-inspired strategies and extending the physics of the particles by evolving, through the use of GP, optimal force generating equations for controlling the particles in a PSO.
Feng et al. \cite{feng2006identification} propose a new method for the simultaneous establishment of a visco-elastic rock material model structure and the related parameters based on a hybrid genetic programming system with an improved particle swarm optimization algorithm.
More specifically, GP explores the model's structure, and a modified version of PSO identifies the parameters in the provisional model.
Besides, Kanemasa and Aiyoshi \cite{kanemasa2014algorithm} adopt PSO as a heuristic optimization method, and they augment PSO by using GP as a meta-algorithm to solve the learning problem of automatically generating tuning rules for the parameters in the PSO algorithm.

\subsection{Surrogate Modeling}
\label{sec:surrogate}

To overcome the difficulty arising from finding the global optimum in rugged fitness landscapes (i.e., those with a lot of local optima), a possibility is the use of a \emph{surrogate model (SM)}  \cite{bhosekar2018advances}. 
The broad consensus on surrogate-based evolutionary frameworks is that the algorithm efficiency can be improved by replacing, as much as possible, costly objective functions with surrogates that are considerably less expensive to build and compute.
In this manner, the overall computational burden of the evolutionary search can be significantly reduced, since the effort required to build and use the surrogates is much lower than in the traditional approach that directly couples the evolutionary algorithm with the costly objective functions~\cite{emmerich2002metamodel,jin2005comprehensive,ong2005surrogate,tenne2009model,ulmer2003evolution}.

Constructing an SM has proven to be very efficient, and SMs are employed in many scientific fields in which evaluating the original fitness function would represent a major obstacle. For instance, SMs have been applied to rotor blade design and optimization \cite{booker1998optimization}, high-speed civil transport \cite{knill1999response}, airfoil shape optimization \cite{rai2000improving}, and diffuser shape optimization \cite{madsen2000response}.

Concerning fitness landscape analysis, the early approaches focused on building \emph{global} surrogates (i.e., surrogates defined on the whole original domain) \cite{ulmer2003evolution} with the scope of modeling the entire fitness landscape. 
Nonetheless, because of the curse of dimensionality \cite{donoho2000high}, several contributions started to tackle the problem using \emph{local} surrogate models (i.e., surrogates defined on a restriction of the original domain) \cite{giannakoglou2002design}, \cite{ong2003evolutionary}, \cite{manzoni2020surfing} or a combination of global and local surrogate models \cite{zhou2006combining}, \cite{zhou2007memetic}.  

Lian et al. \cite{lian2004enhanced} proposed an enhancement for standard GAs using a local surrogate search to expedite their convergence. The model uses a GA to generate a population of individuals and rank them with a real function. Then, a gradient-based local search is performed on the SM to find new, promising solutions. Both GA and local search are alternately used under a trust-region framework until the optimum is found. 

SMs have also been exploited in the context of GP. 
For example, in \cite{kattan2012evolving}, the authors propose an SM based on genetic programming and radial basis function networks (RBFN)  called \emph{GP-RBFN Surrogate}. In particular, they use GP as an optimization engine to evolve both the structure of a radial basis function (RBF) and its parameters to obtain an upgrade of the standard RBFN surrogate model. 
Specifically, GP receives an RBF as a primitive and tries to evolve a new RBF expression with improved width parameters, then the evolved RBFN is used as a surrogate model to improve the GP search.

\section{Material and Methods}
\label{sec: matandmeth}

This section introduces the tools that will be exploited for the development of our method: First, genetic programming is presented in Section~\ref{subsec: geneticprogramming}; subsequently, Fuzzy Self-Tuning Particle Swarm Optimization, a variant of another bio-inspired algorithm, is described in Section~\ref{subsec: fst-pso}. 
Afterward, taking advantage of these methods, a new surrogate modeling technique for dealing with challenging optimization problems is introduced in Section~\ref{subsec: gp-fst-pso}.

\subsection{Genetic Programming}
\label{subsec: geneticprogramming}

Genetic algorithms (GAs), introduced by John Holland in $1975$~\cite{holland1992adaptation}, are a technique for solving both constrained and unconstrained optimization problems based on a natural selection process that mimics biological evolution. The main idea at the base of these methods is to encode and iteratively improve a \emph{population} of individuals that encode candidate solutions (of fixed-length) for a given problem. At each iteration, called a \emph{generation}, candidate solutions (called \emph{parents}) are selected and mutated in order to generate the components (called \emph{offsprings}) of the subsequent generation.

For many problems in machine learning and artificial intelligence, the most natural representation for a solution is a computer program of indeterminate size and shape, as opposed to strings whose size has been determined in advance. 
Genetic programming (GP) overcomes the main limitation of GAs concerning the fixed length of the solutions by providing a method for finding a computer program of unspecified size and shape to solve (or approximately solve) a problem \cite{koza1994genetic}. 
It starts with an initial population of unfit randomly generated programs (i.e., \emph{individuals}), and each of them is evaluated in terms of how well it performs through the introduction of a specific objective function (called the \emph{fitness function}), whose nature varies according to the task of the problem under consideration. 
Some individuals in the population will turn out to be fitter than others, and the Darwinian principle of survival of the fittest (\emph{elitism}) and \emph{genetic operators} are used to create a new (and improved) offspring population starting from the current population of programs.
The process is repeated for a given number of iterations, and, at each step, it will produce populations that exhibit an improvement in the average fitness, which corresponds automatically to an improvement in the quality of the solution. 
The two genetic operators involved in the GP evolutionary process are:

\subsubsection{Crossover}
\label{section crossover}

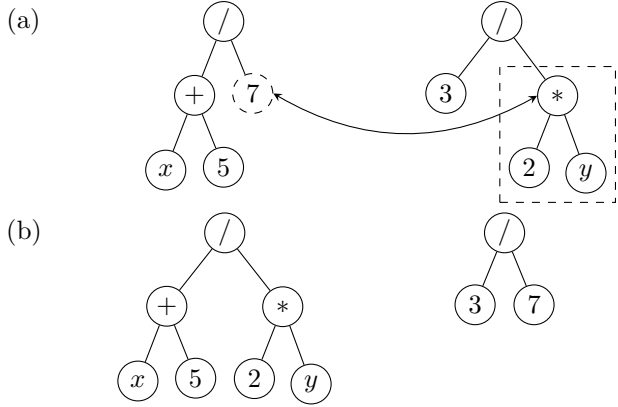
\begin{figure}[h!]

\begin{forest}
for tree={circle, draw, minimum size=1.5em, inner sep=0.5mm}
[,phantom,s sep=2cm
[(a),draw=none]
[$/$, 
  [$+$
    [$x$]
    [$5$]
  ]
   [$7$,draw,dashed,alias=7]
]
[$/$
  [$3$]
  [,phantom]
  [$*$,alias=*,tikz={\node [draw,dashed,fit=() (!1) (!l)] {};}
    [$2$]
    [$y$]
  ]
]
]
\draw [stealth-stealth](7.east) to[bend right] (*.west);
\end{forest}

\begin{forest}
for tree={circle, draw, minimum size=1.5em, inner sep=0.5mm}
[,phantom,s sep=2cm
[(b),draw=none]
[$/$
  [$+$
    [$x$]
    [$5$]
  ]
  [,phantom]
  [$*$,alias=*
    [$2$]
    [$y$]
  ]
]
[$/$
  [$3$]
  [$7$,alias=7]
]
]
\end{forest}
    \caption{Example of GP crossover. Here the selected parents are $\frac{x+5}{7}$ and $\frac{3}{2y}$. In part (a), their trees are cut at some randomly chosen position. Then, in (b), the subtrees are swapped to obtain $\frac{x+5}{2y}$ and $\frac{3}{7}$.}
    \label{fig:crossover}
\end{figure}

The crossover operator induces the mixing of genetic material from two selected parents to produce two new offsprings: once the parents are chosen, the trees that encode the solutions that they represent are cut at some random position, and the subtrees generated through this cut are swapped between them, leading to the creation of two new individuals. An example of this operator is shown in Figure~\ref{fig:crossover}.

\subsubsection{Mutation}

\begin{figure}[h!]
    \centering
\begin{forest}
for tree={circle, draw, minimum size=1.5em, inner sep=0.5mm}
[,phantom,s sep=3cm
[$/$
  [$+$
    [$x$]
    [$5$]
  ]
   [,phantom]
  [$*$,alias=*,tikz={\node [draw,dashed,fit=() (!1) (!l)] {};}
    [$2$]
    [$y$]
  ]
]
[$/$
  [$+$
    [$x$]
    [$5$]
  ]
  [,phantom]
  [$8$,draw,dashed,alias=8]
]
]
\draw [-stealth](*) to[bend right] (8.west);
\end{forest}
    \caption{Example of GP mutation. The program $\frac{x+5}{2y}$ is modified into $\frac{x+5}{8}$.}
    \label{fig:mutation}
\end{figure}
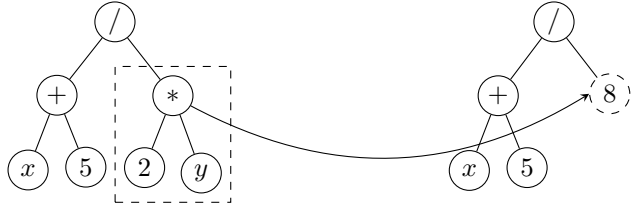

Mutation is a genetic operator which is applied to a single individual: a random number in the interval $[0,1]$ is generated with uniform probability and compared to a predetermined \emph{mutation rate}. If the random number is greater than the mutation rate, no mutation is applied to the individual under consideration; otherwise, the program will be modified.
If this is the case, a random point in the tree is chosen, and subsequently the subtree below this point is replaced by a new, randomly generated subtree.
An example of the mutation operator is shown in Figure~\ref{fig:mutation}. 

\subsection{Fuzzy Self-Tuning Particle Swarm Optimization (FST-PSO)}
\label{subsec: fst-pso}

Particle swarm optimization is a bio-inspired algorithm that performs the task of searching for an optimal solution over a fixed solution space; it was proposed by Kennedy and Eberhart in $1995$ \cite{kennedy1995particle} and was inspired by the motion of birds flocks and schooling fish.
In this algorithm, $P$ particles (i.e., candidate solutions) are placed in a bounded $D$-dimensional search space, and all the particles cooperate to recognize and converge to the (global) optimal solution of the optimization problem.
In the standard formulation of PSO, the movements of the particles are guided by their own best-known position over the search-space (cognitive factor, $c_{cog} \in \mathbb{R}^+$) as well as the entire swarm's best-known position (social factor, $c_{soc} \in \mathbb{R}^+$): when better positions are discovered, these are subsequently used to guide the subsequent movements of the swarm. 
This process is repeated for a predetermined number of epochs, eventually leading the swarm (as a whole) to move closer to the optimum of the fitness function.

Due to their relevance, the coefficients $c_{cog}$ and $c_{soc}$ together with the values of the maximum and minimum velocity ($\Vec{v}_{max},\Vec{v}_{min} \in \mathbb{R}^D$) of particles along each dimension of the search space, which are other important hyperparameters, are usually carefully selected according to the characteristics of the specific problem under consideration.
As it is not possible to determine such values analytically, their selection commonly involves extensive trials. 

To overcome this issue, a variant of the classic PSO framework, named \emph{Fuzzy Self-Tuning PSO} (FST-PSO) that does not require any user settings has been introduced in \cite{nobile2018fuzzy}.
Indeed, by adding a Fuzzy Rule-Based System (FRBS) to each particle, FST-PSO permits a during-the-optimization and independent adjustment of each particle according to its performance and its distance from the position of the current global best particle. 
FRBSs are models based on the concepts of \emph{fuzzy sets} (also known as \emph{uncertainty sets}, as their elements are characterized by a \emph{degree of membership}) proposed by Zadeh in \cite{zadeh1996fuzzy}.
FRBSs are mainly used to address complex real-world problems. 
In particular, FRBSs allow coping with uncertainty, imprecision, and non-linearity by making a model more flexible to change.
The FST-PSO algorithm can be summarized in four steps (an in-depth description, as well as the pseudo-code of the method, can be found in \cite{nobile2018fuzzy}).

The starting point consists in randomly placing all the particles in the search space.
Afterward, as in standard PSO, it proceeds by evaluating the fitness function for all the particles and updating the information about the current global and local bests.
At this point, FST-PSO exploits FRBS to perform multiple Sugeno inferences, proposed by Takagi and Sugeno in $1985$ \cite{takagi1985fuzzy} for developing a systematic approach to generating fuzzy rules from a given input-output dataset. 
A typical fuzzy rule in a Sugeno fuzzy model has the form:
    \begin{center}
        if $x$ is $A$ and $y$ is $B$ then $z=f(x,y)$,
    \end{center}
where $A$ and $B$ are fuzzy sets, and $f(x, y)$ is usually a polynomial in the input variables $x$ and $y$.
In the case of FST-PSO, the output polynomial depends on each particle’s performance with respect to the previous iteration and its distance from the current global best in the swarm.
Finally, the velocity and position of all particles are updated, and the algorithm repeats until the exhaustion of the fitness evaluations' budget and, as the result of the optimization, the best solution found by the swarm is returned.

\subsection{GP-FST-PSO Surrogate Model}
\label{subsec: gp-fst-pso}

Consider a $D-$dimensional search space $S \subseteq \mathbb{R}^D$ and a function
$f:\mathbb{R}^D \rightarrow \mathbb{R}$. A \emph{minimization} (maximization) problem consists in finding the optimal $\Bar{x} \in \mathbb{R}^D$ such that $f(\Bar{x}) \le f(x)$ $(f(\Bar{x}) \ge f(x))$ $\forall x \in \mathbb{R}^D \setminus \{ \Bar{x} \}$. 
Therefore, the function $f$ associates a correspondent measure to a candidate solution in the search space, inducing a hyper-surface $\Lambda$, called a \emph{fitness landscape}, representing the quality of all the possible solutions in the search space. 
Especially when dealing with applications related to real-life problems, this landscape appears irregular and noisy and is also often characterized by the existence of a considerable number of local optima, which can lead the optimizer to premature convergence. 

To deal with this kind of scenario, we combine the two techniques mentioned above, i.e. GP (\ref{subsec: geneticprogramming}) and FST-PSO (\ref{subsec: fst-pso}).
The goal of the proposed method, named the GP-FST-PSO Surrogate Model (which in turn stands for \emph{Genetic Programming Fuzzy Self-Tuning Particle Swarm Optimization Surrogate Model}), is the definition of a smooth function, easier to optimize with respect to the original one, that still maintains a fundamental property of the original rugged fitness landscape: the location of the argminimum.

To accomplish the creation of such a surrogate model, we consider a genetic programming approach: the surrogate function will be created by starting from an initial random population of functions and then applying genetic operators (mutation and crossover) to this population for a predetermined number of generations.
The FST-PSO algorithm enters the picture for the definition of the loss function $\mathcal{L}$. 
The loss function that evaluates the quality of a solution must implicitly reward solutions characterized by the same argminimum as the original fitness landscape, as the aim of an optimization problem consists in finding the coordinates where the function is minimized. 
Therefore, the loss utilized in this paper is computed as follows: when a surrogate function is created via the GP routine, its argminimum is computed through the FST-PSO algorithm. 
Such computation does not involve a considerable computational effort, on the contrary, finding the argminimum of GP-generated functions is a computationally easy task. Indeed, the population consists of smooth functions, as the only operators used for their definitions are mathematical elementary operators (summation, subtraction, multiplication, and division) and two other typical GP operators (SWAP and DUP) that, in any case, do not lead to any noise in the construction of a function. 
Once the argminimum is computed, it is passed as the argument of the benchmark noisy function and the result is used as a measure for defining the loss of the GP-FST-PSO Surrogate Modeling method. The loss $\mathcal{L}$, therefore, can be defined as follow:
\begin{equation}
    \mathcal{L} = f (\text{FST-PSO} (\Tilde{f})),
\end{equation}
where $\Tilde{f}$ is a function generated through GP and $f$ is the original benchmark function. 
This quantity is minimized with respect to all the individuals of the current generation, and the best result leads to the best individual of the generation under consideration. 
This choice ensures that the functions whose argminimum is near to the one of the original function are rewarded and survive through epochs. Finally, when the argminimum of the surrogate function reaches the argminimum of the original noisy benchmark function, the loss reaches its minimum as the original benchmark function cannot return any lower values. 
The use of an evolutionary method, such as GP, prevents the search from falling into local minima and getting stuck there, as it allows big jumps in the solution space. In other words, at each step, new areas of this space can be explored and discovered.

As the goal of the proposed method consists not only in finding the correct location in the search space where the argminimum lies but also in approximating the fitness landscape's shape, the aforementioned loss considers an additional term.
This term consists of the root mean squared error (RMSE) computed between the surrogate function and the original benchmark function over a set of points uniformly sampled from the search space.
The RMSE is added to the aforementioned loss $\mathcal{L}$ and the whole quantity is minimized. Thus, both the correctness of the argminimum location and the fact that the surrogate function consistently represents the original one are considered. The final loss can be outlined as follows:
\begin{equation}
    \mathcal{L} = f (\text{FST-PSO} (\Tilde{f})) + \text{RMSE}(f, \Tilde{f}).
\end{equation}
The algorithm that summarizes the aforementioned proposed optimization method is presented in Algorithm~\ref{alg:alg1}.

\begin{algorithm}
    \begin{algorithmic}[1]
        \Function{GP-FST-PSO}{$f$, search\_space, pop\_sz, n\_gen}
            \State sampled\_points $\gets$ \textbf{sample}($f$, search\_space)\;
            \State pop $\gets$ \textbf{random\_pop}(pop\_sz)\;
            \For{$1 \le n \le$ n\_gen} 
                \State offsp $\gets$ \textbf{tournament}(pop, FST-PSO)\; 
                \Comment{crossover}
                \State pop $\gets$ \textbf{mutation}(offsp)\; \Comment{mutation}
                \State pop $\gets \min_{\Tilde{f} \in \text{offsp}} \text{FST-PSO}(\Tilde{f}) $\; \Comment{elitism} 
            
            \EndFor
            \State best $\gets \min_{\Tilde{f} \in \text{pop}} \text{FST-PSO}(\Tilde{f}) $\; 
            \State \Return best\;
        \EndFunction
    \end{algorithmic}
    \caption{\label{alg:pseudocode}GP-FST-PSO Surrogate Model pseudocode, where $f$ is the benchmark function that must be optimized, search\_space is the search space domain relative to $f$, pop\_sz is the size of the population, and n\_gen is the number of generations considered.}
    \label{alg:alg1}
\end{algorithm}

\section{Experimental Settings}
\label{sec: expsett}
This section describes the benchmark functions used to validate the method (Sec. \ref{subsec: benchmark}) and then presents the experimental setting (Sec. \ref{subsec: exp-setting}) needed to make the experiments completely reproducible. 
The code, for complete reproducibility of the proposed experiments, is available at \url{https://github.com/gpietrop/SurMod-GP-FST-PSO} \cite{code}.

\subsection{Benchmark Functions}
\label{subsec: benchmark}

\begin{table}
\centering
\begin{tabular}{|c|ll|}
\hline 
{Ackley}   & \multicolumn{2}{l|}{$f_{Ack}(\vec{x})=20 + e - 20 \text{exp}(-0.2 \sqrt{\frac{1}{D}\sum_{d=1}^D d_d^2})-$} \\  [1ex]
                    & \multicolumn{2}{l|}{$\exp \left(\frac{1}{D} \sum_{d=1}^D|x_d \sin(x_d) + 0.1x_d| \right)$} \\ 
                    & \multicolumn{1}{l}{Search space}                          & $[-30,30]^D$                                 \\ [1ex]
                    & \multicolumn{1}{l}{Minimum}                               & $f_{Ack}(\vec{0})=0$                         \\ [1ex] \hline
{Alpine}   & \multicolumn{2}{l|}{$f_{Alp}(\vec{x})=\sum_{d=1}^D |x_d \sin(x_d) + 0.1x_d |$} \\ [1ex] 
                    & \multicolumn{1}{l}{Search space}                          & $[-10,10]^D$                                 \\ [1ex]
                    & \multicolumn{1}{l}{Minimum}                               & $f_{Alp}(\vec{0})=0$                         \\ [1ex] \hline
{Griewank} & \multicolumn{2}{l|}{$f_{Gri}(\vec{x})=\frac{1}{4000}\sum_{d=1}^D x_d^2 - \prod_{d=1}^D \cos \left( \frac{x_d}{\sqrt{d}} \right) +1$} \\ [1ex] 
                    & \multicolumn{1}{l}{Search space}                          & $[-600,600]^D$                                 \\ [1ex]
                    & \multicolumn{1}{l}{Minimum}                               & $f_{Gri}(\vec{0})=0$                         \\ [1ex] \hline
{Michalewicz}       & \multicolumn{2}{l|}{$f_{Mic}(\vec{x})=- \sum_{d=1}^D \sin(x_d) \sin^{2k} \left( \frac{dx_d^2}{\pi} \right)$,} \\ [1ex]
                    & \multicolumn{2}{l|} {$k=10$ in this paper} \\ [1ex] 
                    & \multicolumn{1}{l}{Search space}                          & $[-0,\pi]^D$                                 \\  [1ex]
                    & \multicolumn{1}{l}{Minimum}                               & $f_{Mic}(2.2044,1.5692)=-1.801$                         \\ [1ex] \hline
{Rastring} & \multicolumn{2}{l|}{$f_{Ras}(\vec{x})=10D + \sum_{d=1}^D(x_d^2 - 10\cos(2\pi x_d))$} \\ [1ex] 
                    & \multicolumn{1}{l}{Search space}                          & $[-5.12,5.12]^D$                                 \\ [1ex]
                    & \multicolumn{1}{l}{Minimum}                               & $f_{Ras}(\vec{0})=0$                         \\ [1ex] \hline
{Rosenbrock}       & \multicolumn{2}{l|}{$f_{Ros}(\vec{x})=\sum_{d=1}^{D-1}[10(x_d^2 - x_{d+1})^2 + (x_d-1)^2]$} \\ [1ex] 
                    & \multicolumn{1}{l}{Search space}                          & $[-5,10]^D$                                 \\ [1ex]
                    & \multicolumn{1}{l}{Minimum}                               & $f_{Ros}(\vec{1})=0$                         \\ [1ex] \hline
{Schwefel}        & \multicolumn{2}{l|}{$f_{Sch}(\vec{x})=418.9829D - \sum_{d=1}^D x_d \sin(\sqrt{|x_d|})$} \\ [1ex] 
                    & \multicolumn{1}{l}{Search space}                          & $[-500,500]^D$                                 \\ [1ex]
                    & \multicolumn{1}{l}{Minimum}                               & $f_{Sch}(\overrightarrow{420.9687})=0$           \\ [1ex] \hline
{Vincent}          & \multicolumn{2}{l|}{$f_{Vin}(\vec{x})=\sum_{d=1}^D \sin \left(10 \log(x_d)\right)$} \\ [1ex] 
                    & \multicolumn{1}{l}{Search space}                          & $[0.25,10]^D$                                 \\ [1ex]
                    & \multicolumn{1}{l}{Minimum}                               & $f_{Vin}(\overrightarrow{0.25})=-D$                         \\ [1ex] \hline
{Xin-She } & \multicolumn{2}{l|}{$f_{Xin}(\vec{x})=\sum_{d=1}^D |x_d| [\exp\left(\sum_{d=1}^D \sin(x_d^2) \right)]^{-1}$} \\ [1ex] 
      Yang n.2      & \multicolumn{1}{l}{Search space}                          & $[-2\pi,2\pi]^D$                                 \\ [1ex]
                    & \multicolumn{1}{l}{Minimum}                               & $f_{Xin}(\vec{0})=0$                         \\ [1ex] \hline
\end{tabular}
\caption{Closed expressions of the considered benchmark functions together with the domain, the location of the minimum and the global minimum value.}
\label{tab:bechmark_fun}
\end{table}

The set of functions used to assess the proposed method is listed in Table \ref{tab:bechmark_fun}.

Here we summarize the main characteristics of the functions, such as their closed-form expression, their domain, and the value of their global minimum, together with the correspondent argminimum.
These functions (all widely used as benchmarks problems in the optimization framework \cite{jamil2013literature}, \cite{liu2019ai}) belong to a subset of the \emph{CEC $2005$} test suite. They were selected because their structural characteristics of noise and ruggedness make them excellent candidates for studying the effectiveness of the proposed approach.
The considered benchmark functions (difficult to optimize, non-convex, and multimodal) are Ackley, Alpine, Griewank, Michalewicz, Rastring, Rosenbrock, Schwefel, Vincent, and Xin-She Yang 2; all of them are characterized by various challenges and intrinsic difficulty. 
For example, Ackley has the global optimum located in a very small basin \cite{plevris2022collection}, Griewank is characterized by a number of local minima that increases exponentially according to the dimension $d$ \cite{plevris2022collection}, Michalewicz presents $d!$ local minima \cite{molga2005test}, and Vincent has $6^D$ global optima with vastly different spacing between them, while it does not present local optima \cite{li2013benchmark}.
In Rosenbrock, the global minimum is inside a long, narrow, parabolic-shaped flat valley: to find the valley is trivial, but to converge to the global minimum is difficult \cite{plevris2022collection}. 
Moreover, all but the Rosenbrock function contain at least one trigonometric term, which provided a good way of imitating noise by introducing several local minima in the landscape.
Regarding the Michalewicz function, we only reported its argminimum location and the respective minimum for $D=2$, the typical dimensionality considered in the literature for this benchmark function.

\subsection{Experimental Study}
\label{subsec: exp-setting}

To make the experiments fully reproducible, this section describes the experimental settings (also reported in Table \ref{tab:exp settings}).

\begin{table}[H]
\normalsize
\centering
\begin{tabular}{r|r}
\multicolumn{1}{c|}{\textbf{Parameter}} & \multicolumn{1}{c}{\textbf{Value}}                             \\ \hline
Functions Set                   & +, -, *, //, DUP, SWAP \\
Number of runs                  & 30                    \\
First gen. individuals length   & 10                    \\
Population size ($D=2, 3$)      & 50                    \\
Population size ($D=4$)         & 100                    \\
Number of generations           & 100                   \\
Mutation rate                   & 0.2                   \\
Selection method                & Tournament of size 4  \\
Elitism                         & Best individuals survive 
\end{tabular}
\caption{Experimental settings. In the function set, DUP duplicates the element of a stack, SWAP swaps two elements at top of the stack.}
\label{tab:exp settings}
\end{table}

For each benchmark function, $30$ runs were performed to obtain statistically robust results.
The initial population consists of $50$ randomly generated individuals for problem dimensionality equal to $2$ or $3$, while it is equal to $100$ for problem dimensionality equal to $4$ (a larger number of individuals is chosen for higher dimensionality as the search for the minimum becomes more difficult when the dimensionality of the problem increases).
Each initial individual is a random program of length $10$, generated with the ramped half and half technique.
Only the functions belonging to the function set introduced in Table~\ref{tab:exp settings} are considered, while the constants are selected from different intervals defined according to the domain and codomain boundaries of the benchmarks.

The linear GP exploited for evolving surrogate models is based on a stack architecture (see \cite{perkis1994stack} for a detailed description). 
The algorithm adopted in this work is a \emph{steady state} GP, where a single pair of parents is drawn from the current population at each iteration to produce two children.
We also use an elitist strategy: in each generation, we maintain the individual with the best fitness.
Concerning selection, we rely on a tournament operator in which the best two out of four randomly sampled individuals are selected for the two-points-crossover (see Sec. \ref{section crossover} for details), which is subsequently performed.
The mutation operator replaces, with probability $p_m=0.2$, an instruction (function or constant) with another random function or constant.

To assess the validity of the proposed method across the different benchmark functions, the fitness employed considers the minimum value computed passing the argminimum (computed using FST-PSO) of the surrogate function generated at the last generation of the method into the original benchmark function.
A fitness value correspondent to the real value of the minimum of the benchmark function (reported in Table~\ref{tab:bechmark_fun}) implies that GP-FST-PSO is capable of correctly generating a surrogate model that shares the same argminimum of the original landscape, thus providing a valid and simpler alternative for easily solving the initial optimization task, free of its local optima and consequently of its irregularities and ruggedness.

\section{Results}
\label{sec: results}

\begin{table}[H]
\centering
\begin{tabular}{|ccc|ccc|}
\hline
\multicolumn{3}{|c|}{\textbf{Ackley}} & \multicolumn{3}{c|}{\textbf{Alpine}} \\ \hline
\multicolumn{1}{|c|}{2} & \multicolumn{1}{c|}{3} & 4 & \multicolumn{1}{c|}{2} & \multicolumn{1}{c|}{3} & 4 \\ \hline
\multicolumn{1}{|c|}{1.57e-03} & \multicolumn{1}{c|}{1.01e-02} & 6.91e-02 & \multicolumn{1}{c|}{3.14e-05} & \multicolumn{1}{c|}{2.37e-04} & 7.99e-04 \\ \hline
\multicolumn{3}{|c|}{\textbf{Griewank}} & \multicolumn{3}{c|}{\textbf{Michalewicz}} \\ \hline
\multicolumn{1}{|c|}{2} & \multicolumn{1}{c|}{3} & 4 & \multicolumn{1}{c|}{2} & \multicolumn{1}{c|}{3} & 4 \\ \hline
\multicolumn{1}{|c|}{5.18e-05} & \multicolumn{1}{c|}{2.61e-03} & 5.00e-02 & \multicolumn{1}{c|}{-1.68} & \multicolumn{1}{c|}{-} & - \\ \hline
\multicolumn{3}{|c|}{\textbf{Rastring}} & \multicolumn{3}{c|}{\textbf{Rosenbrock}} \\ \hline
\multicolumn{1}{|c|}{2} & \multicolumn{1}{c|}{3} & 4 & \multicolumn{1}{c|}{2} & \multicolumn{1}{c|}{3} & 4 \\ \hline
\multicolumn{1}{|c|}{1.65e-06} & \multicolumn{1}{c|}{1.47e-04} & 2.89e-03 & \multicolumn{1}{c|}{1.49e-07} & \multicolumn{1}{c|}{6.32e-04} & 4.48e-02\\ \hline
\multicolumn{3}{|c|}{\textbf{Schwefel}} & \multicolumn{3}{c|}{\textbf{Vincent}} \\ \hline
\multicolumn{1}{|c|}{2} & \multicolumn{1}{c|}{3} & 4 & \multicolumn{1}{c|}{2} & \multicolumn{1}{c|}{3} & 4 \\ \hline
\multicolumn{1}{|c|}{9.24e-05} & \multicolumn{1}{c|}{1.80e-01} & 2.46 & \multicolumn{1}{c|}{-1.93} & \multicolumn{1}{c|}{-2.88} & -3.82  \\ \hline
\end{tabular}
\begin{tabular}{|ccc|}
\multicolumn{3}{|c|}{\textbf{Xin-She Yang n.2}} \\ \hline
\multicolumn{1}{|c|}{2} & \multicolumn{1}{c|}{3} & 4 \\ \hline
\multicolumn{1}{|c|}{1.47e-04} & \multicolumn{1}{c|}{1.33e-03} & 2.72e-01 \\ \hline
\end{tabular}
\caption{Median of the fitness values computed over the $30$ runs performed for the considered benchmark functions in dimensions $2$, $3$, and $4$, respectively.}
\label{tab:fitness_values}
\end{table}

\begin{figure*}[!t]


\centerline{\subfloat[]{\includegraphics[width=1.65in]{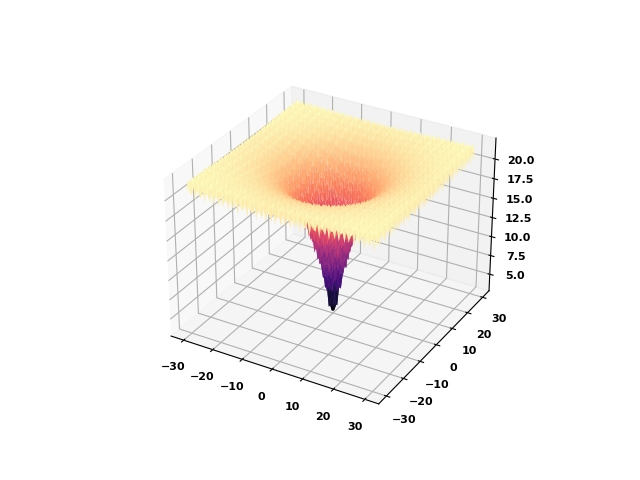}%
{\includegraphics[width=1.65in]{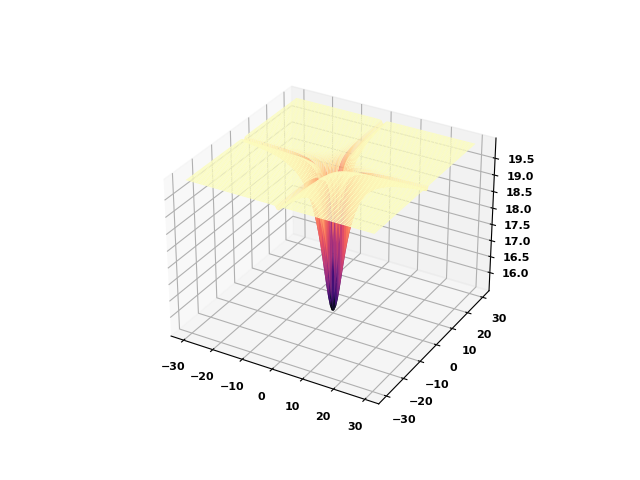}}
\label{fig: ackley_approx}}
\subfloat[]{\includegraphics[width=1.65in]{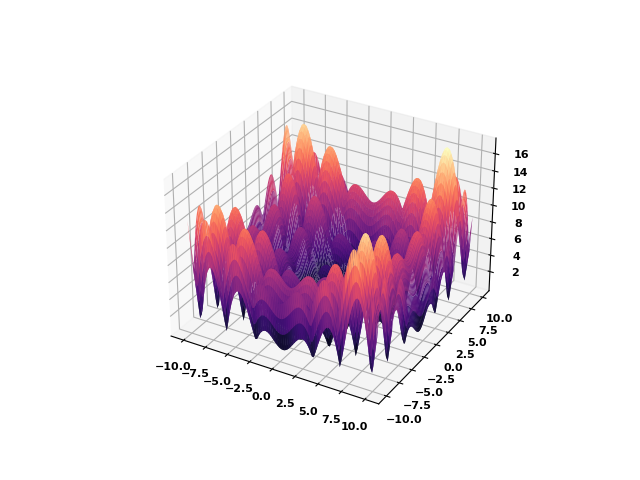}%
{\includegraphics[width=1.65in]{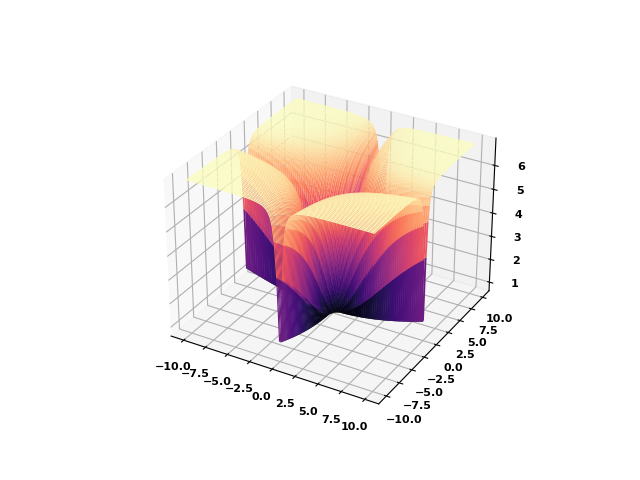}%
\label{fig: alpine_approx}}}}

\centerline{\subfloat[]{\includegraphics[width=1.65in]{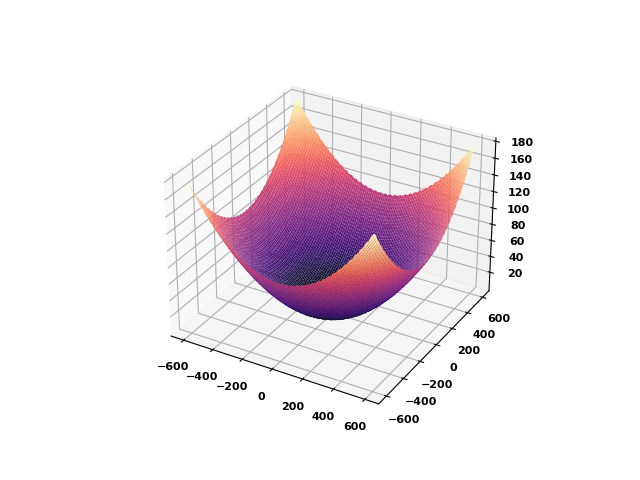}%
{\includegraphics[width=1.65in]{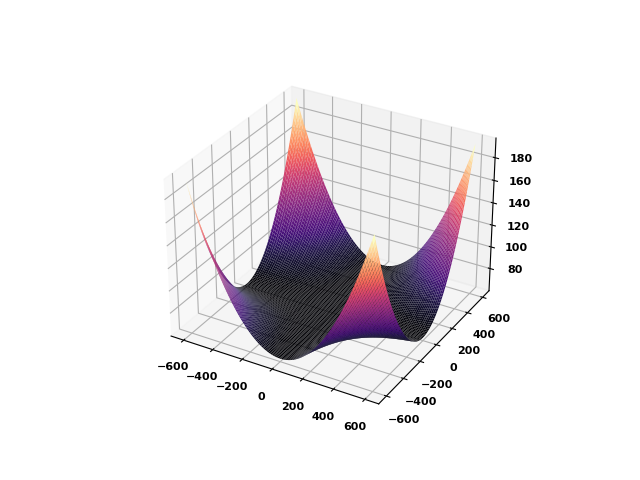}}%
\label{fig: gri_approx}}
\subfloat[]{\includegraphics[width=1.65in]{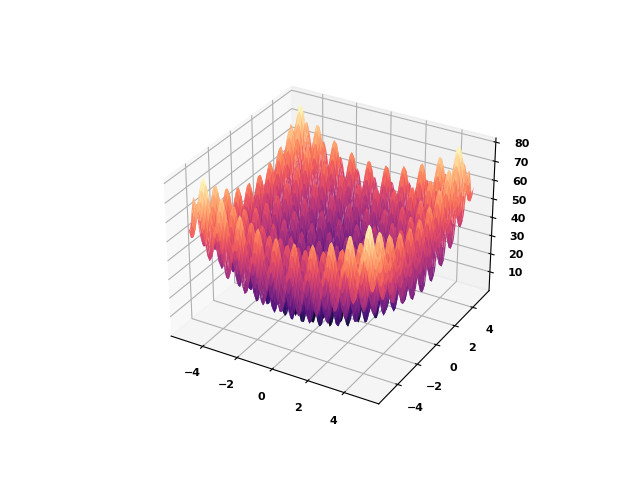}%
{\includegraphics[width=1.65in]{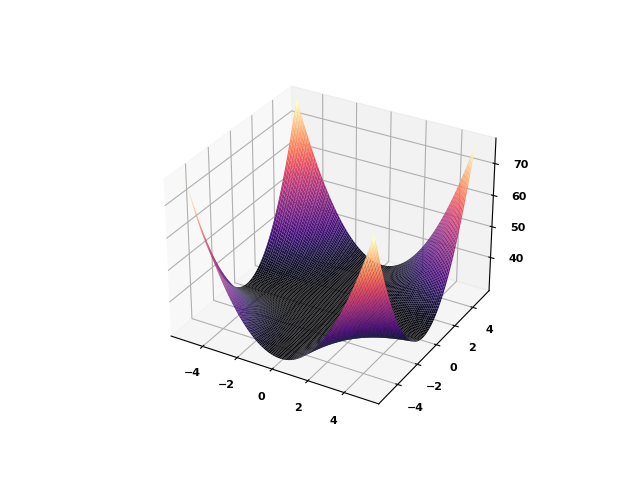}%
\label{fig: rast_approx}}}}

\centerline{\subfloat[]{\includegraphics[width=1.65in]{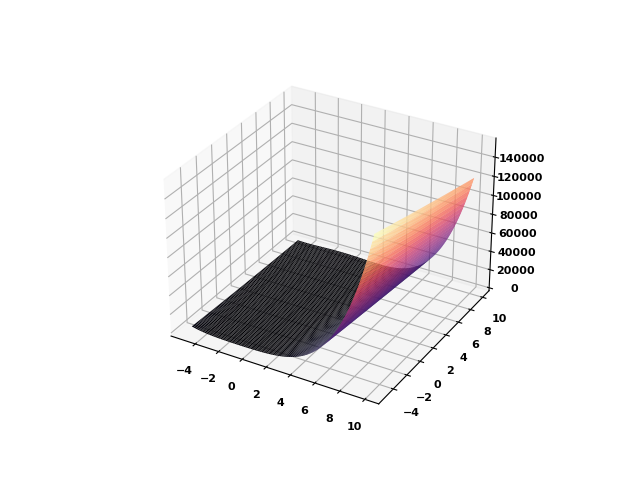}%
{\includegraphics[width=1.65in]{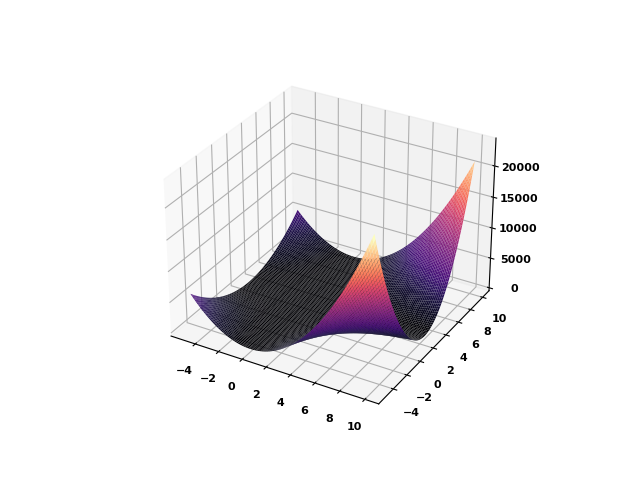}%
\label{fig: xins_approx}}}
\subfloat[]{\includegraphics[width=1.65in]{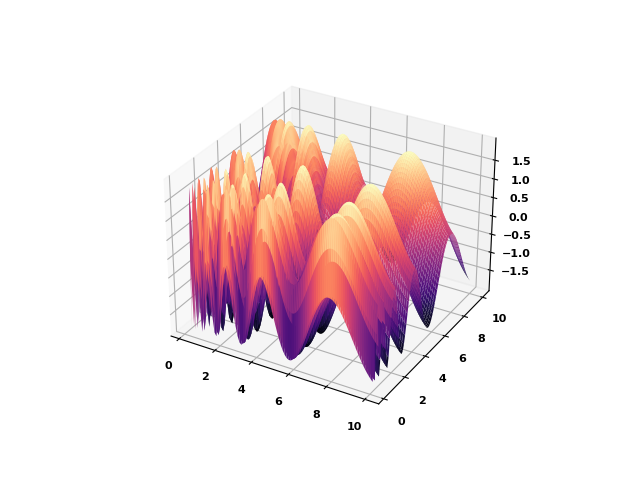}%
{\includegraphics[width=1.65in]{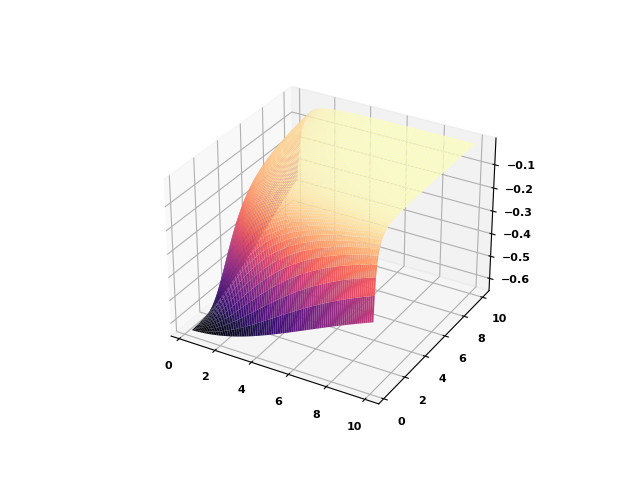}}
\label{fig: vinn_approx}}}

\centerline{\subfloat[]{\includegraphics[width=1.65in]{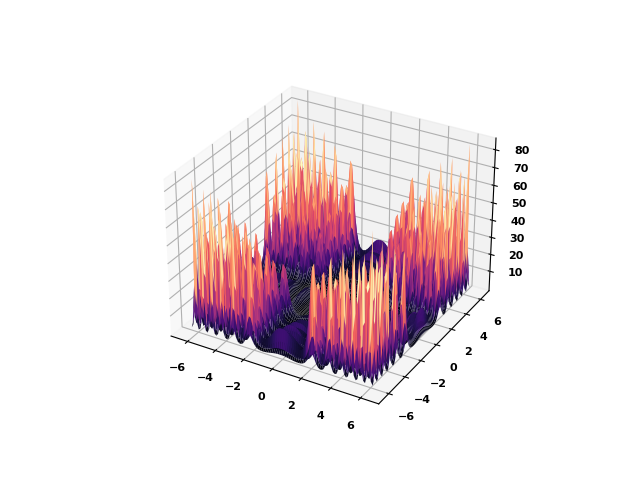}%
{\includegraphics[width=1.65in]{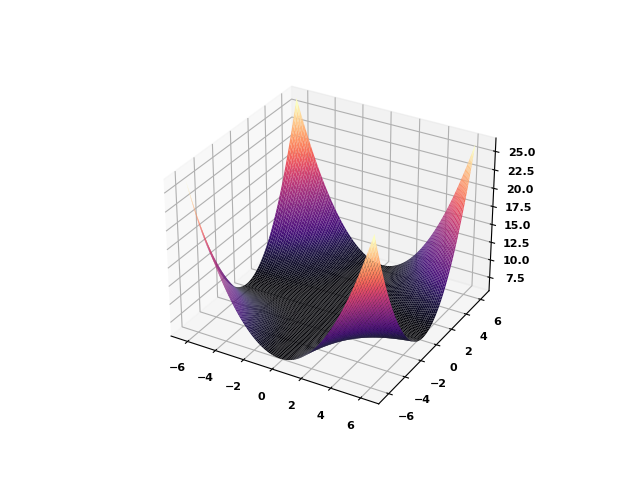}%
\label{fig: xins_approx}}}}

\caption{Example of $D=2$ surrogate models of some of the benchmark functions defined in Table~\ref{tab:bechmark_fun}, created by the \emph{GP-FST-PSO Surrogate Model}. The original fitness landscape (on the left) is compared with the surrogate model (on the right). Benchmark functions considered are: (a) Ackley, (b) Alpine, (c) Griewank, (d) Rastring, (e) Rosenbrock, (f) Vincent, (g) Xin-She Yang n.$2$.}
\label{fig:fun_approx}
\end{figure*}

\begin{figure*}[!t]

\centering
\subfloat[]{\includegraphics[width=1.5in]{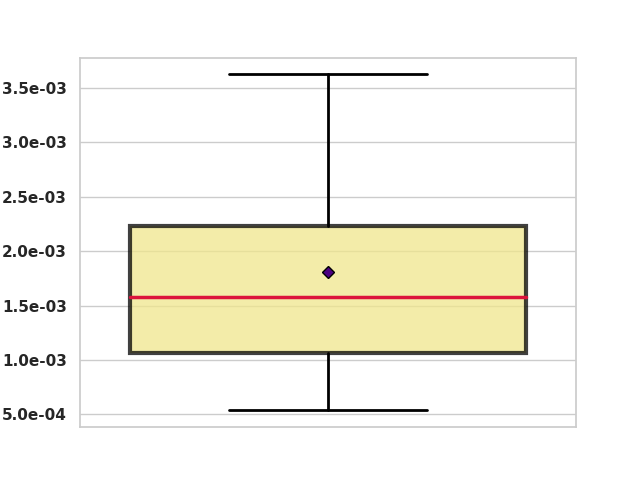}%
\label{fig: ackley 2d_bp}}
\subfloat[]{\includegraphics[width=1.5in]{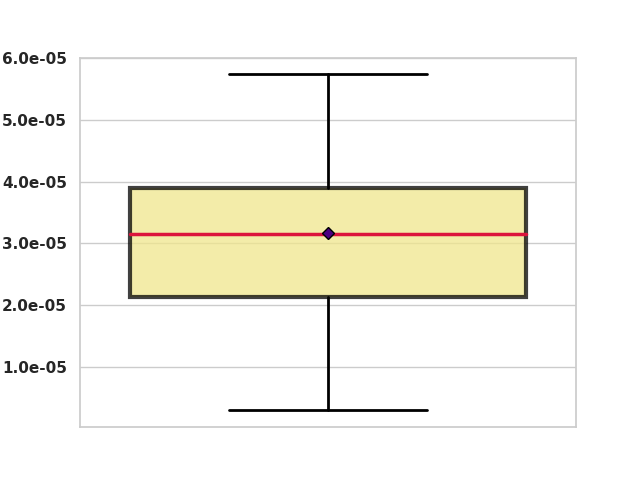}%
\label{fig: alpine 2d_bp}}
\subfloat[]{\includegraphics[width=1.5in]{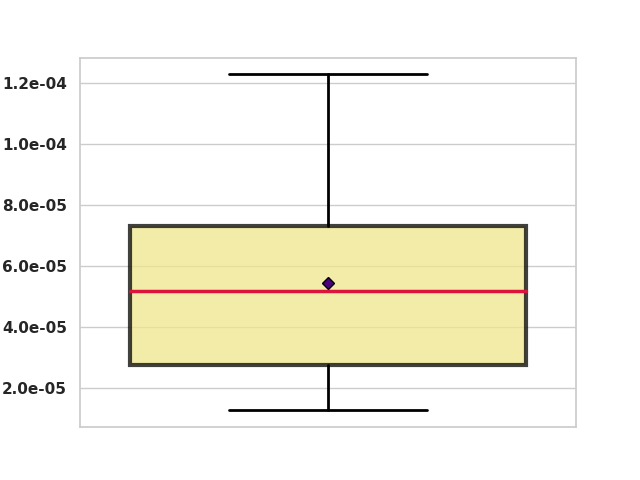}%
\label{fig: grie 2d_bp}}

\subfloat[]{\includegraphics[width=1.5in]{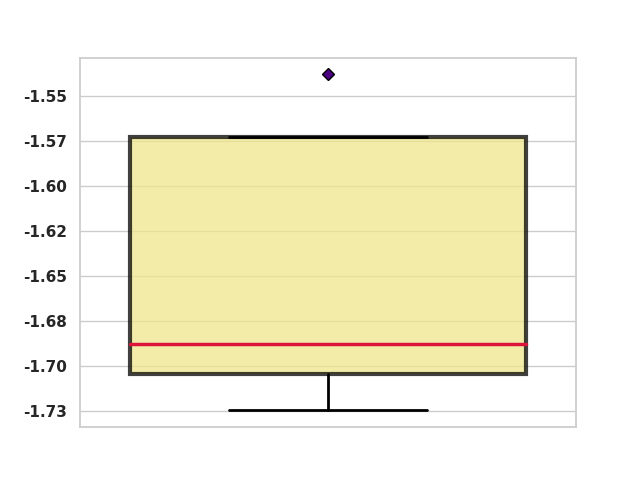}
\label{fig: miche 2d_bp}}
\subfloat[]{\includegraphics[width=1.5in]{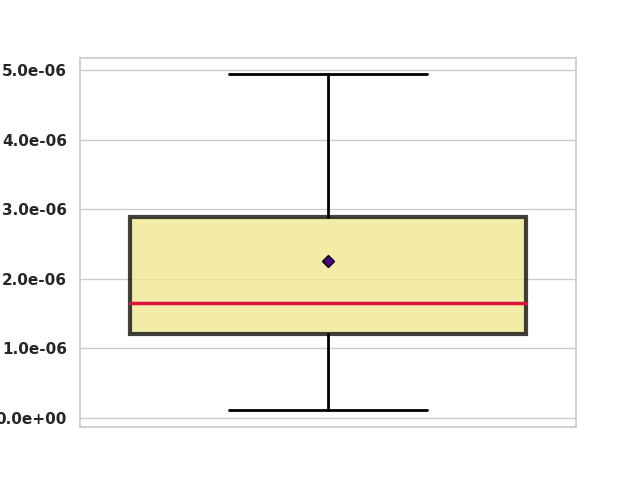}%
\label{fig: rastr 2d_bp}}
\subfloat[]{\includegraphics[width=1.5in]{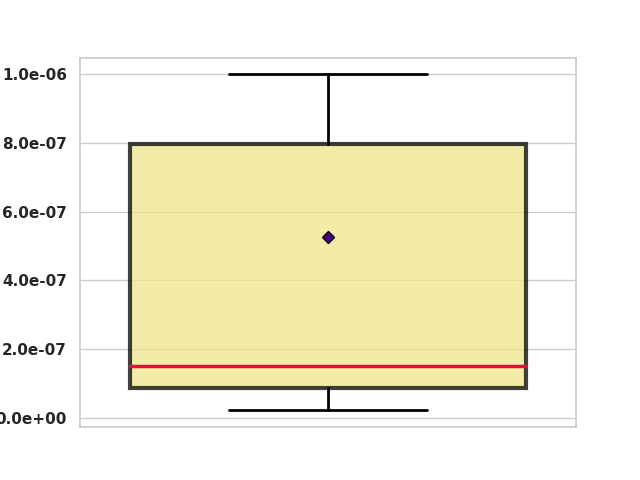}%
\label{fig: rosy 2d_bp}}

\subfloat[]{\includegraphics[width=1.5in]{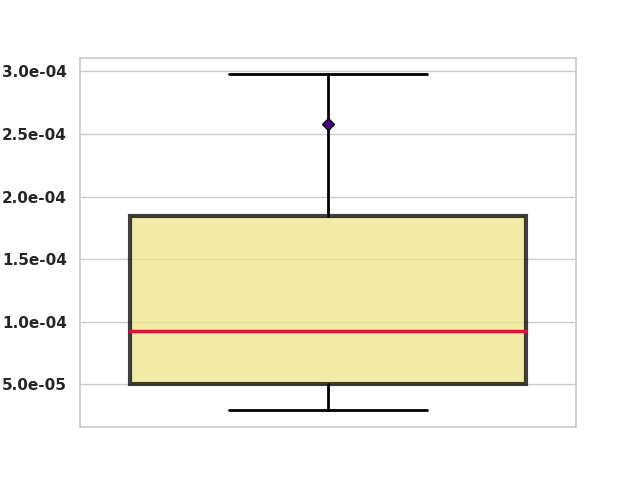}%
\label{fig: schew 2d_bp}}
\subfloat[]{\includegraphics[width=1.5in]{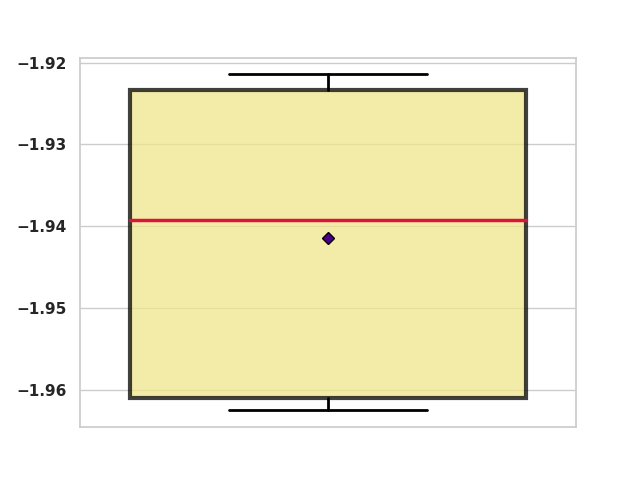}%
\label{fig: vinny 2d_bp}}
\subfloat[]{\includegraphics[width=1.5in]{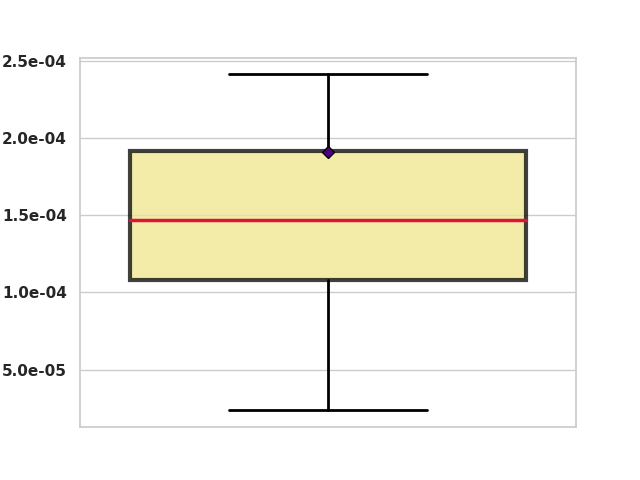}%
\label{fig: xin 2d_bp}}

\caption{Boxplots of fitness distributions over the $30$ independent runs performed for the considered benchmark functions considering $D=2$. The red line represents the median, while the black dot represents the mean. (a) Ackley, (b) Alpine, (c) Griewank, (d) Michalewicz, (e) Rastring, (f) Rosenbrock, (g) Schwefel, (h) Vincent, (i) Xin-She Yang n.2.}
\label{fig:bp_dim2}
\end{figure*}

\begin{figure*}[!t]

\centering
\subfloat[]{\includegraphics[width=1.5in]{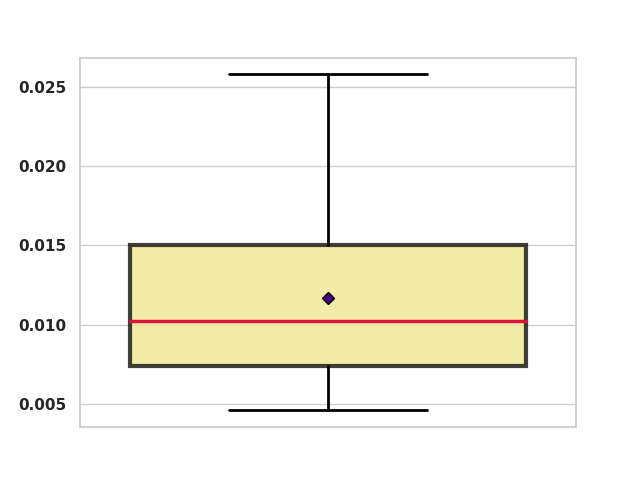}%
\label{fig: ackley 3d_bp}}
\subfloat[]{\includegraphics[width=1.5in]{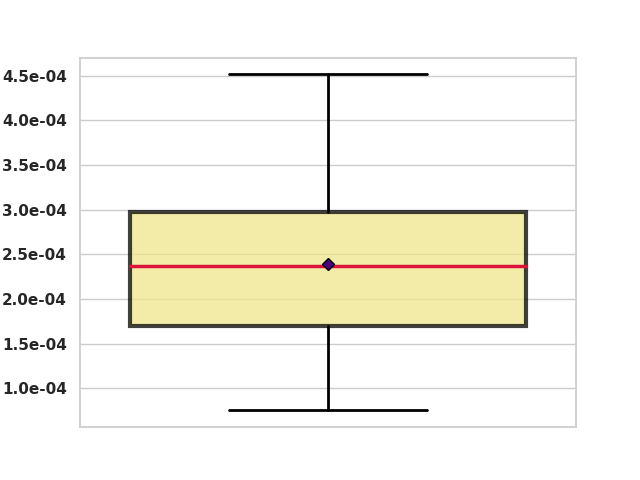}%
\label{fig: alpine 3d_bp}}
\subfloat[]{\includegraphics[width=1.5in]{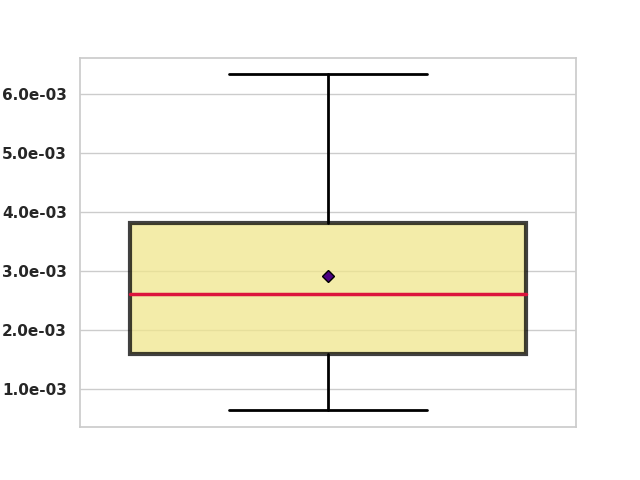}%
\label{fig: grie 3d_bp}}

\subfloat[]{\includegraphics[width=1.5in]{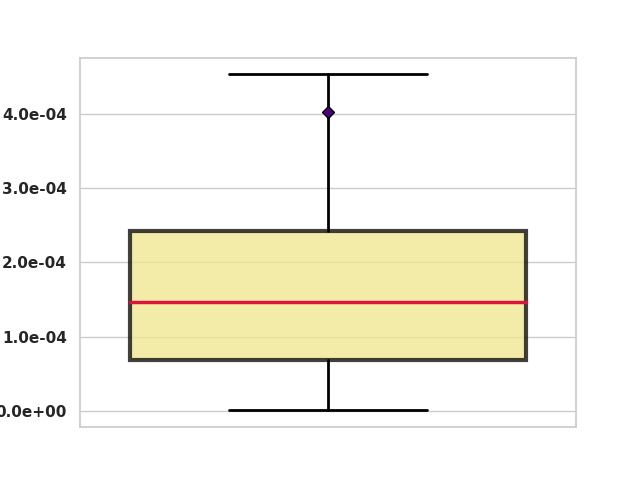}%
\label{fig: rastr 3d_bp}}
\subfloat[]{\includegraphics[width=1.5in]{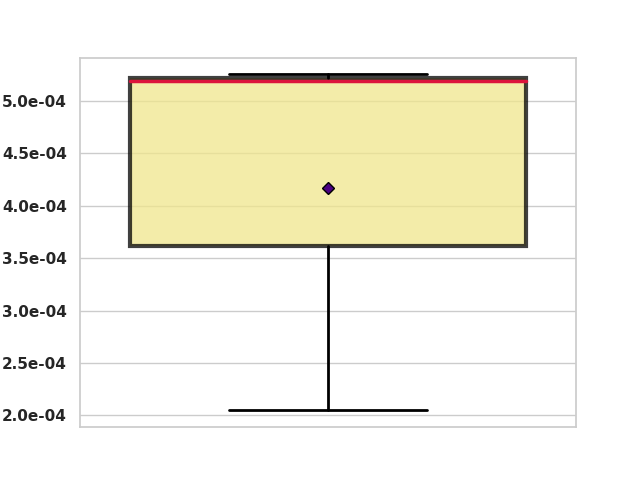}%
\label{fig: rosy 3d_bp}}
\subfloat[]{\includegraphics[width=1.5in]{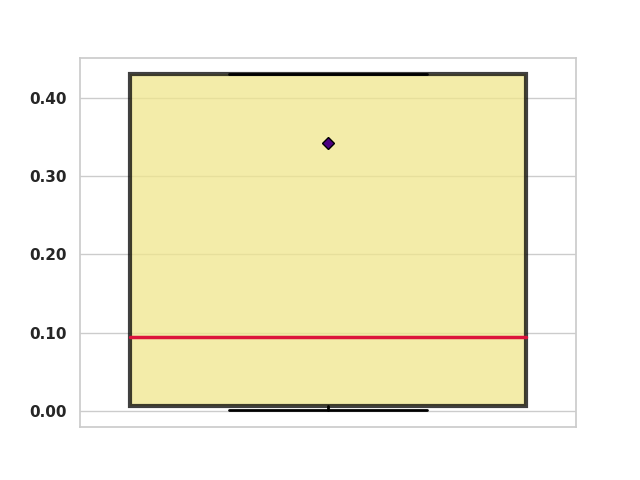}%
\label{fig: schew 3d_bp}}

\subfloat[]{\includegraphics[width=1.5in]{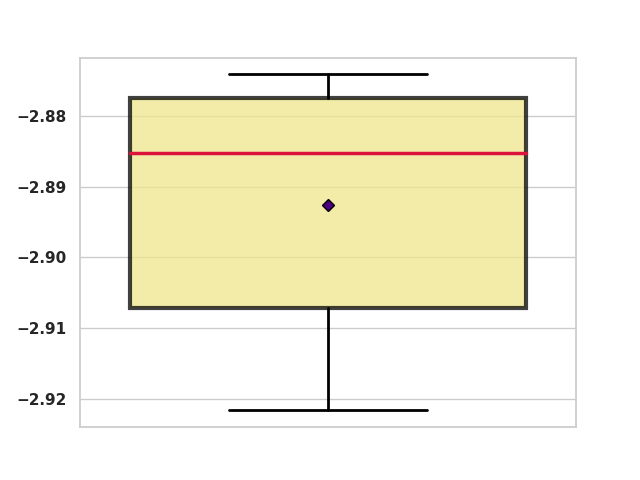}%
\label{fig: vinny 3d_bp}}
\subfloat[]{\includegraphics[width=1.5in]{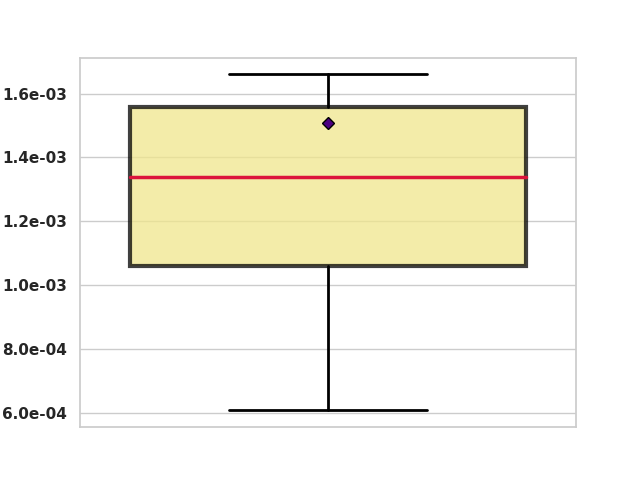}%
\label{fig: xin 3d_bp}}

\caption{Boxplots of fitness distributions over the $30$ independent runs performed for the considered benchmark functions considering $D=3$. The red line represents the median, while the black dot represents the mean. (a) Ackley, (b) Alpine, (c) Griewank, (d) Rastring, (e) Rosenbrock, (f) Schwefel, (g) Vincent, (h) Xin-She Yang n.2.}
\label{fig:bp_dim3}
\end{figure*}

\begin{figure*}[!t]

\centering
\subfloat[]{\includegraphics[width=1.5in]{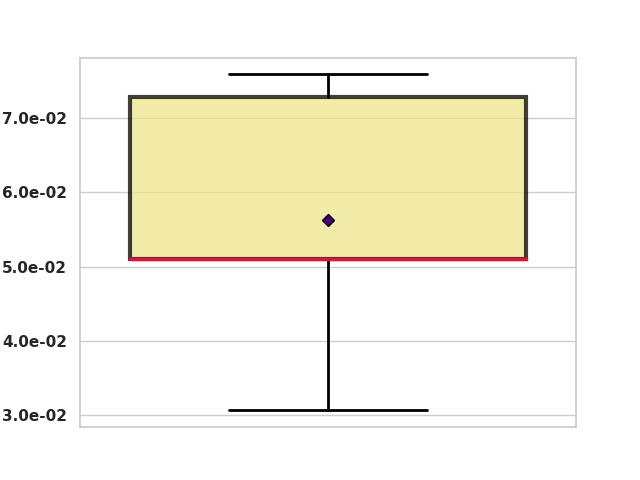}%
\label{fig: ackley 4d_bp}}
\subfloat[]{\includegraphics[width=1.5in]{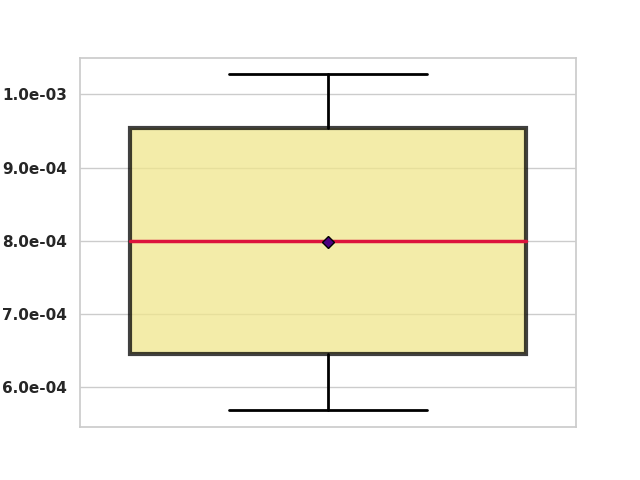}%
\label{fig: alpine 4d_bp}}
\subfloat[]{\includegraphics[width=1.5in]{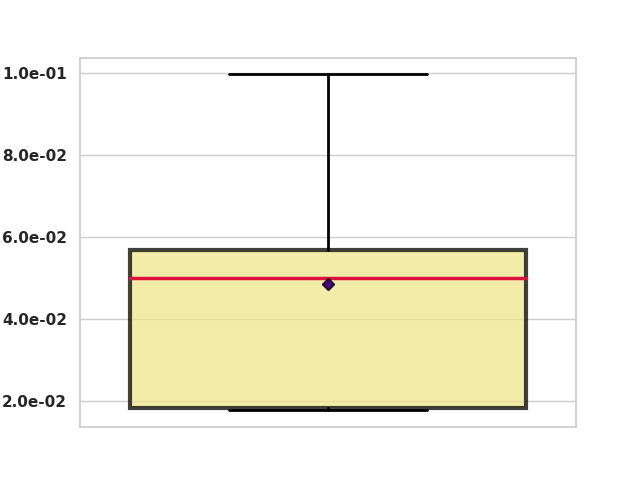}%
\label{fig: grie 4d_bp}}

\subfloat[]{\includegraphics[width=1.5in]{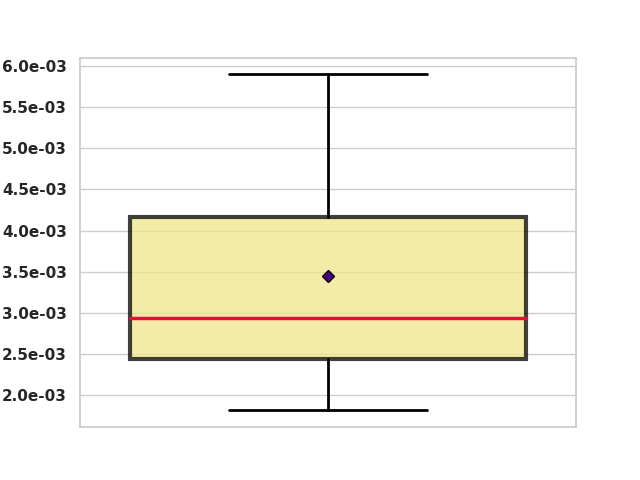}%
\label{fig: rastr 4d_bp}}
\subfloat[]{\includegraphics[width=1.5in]{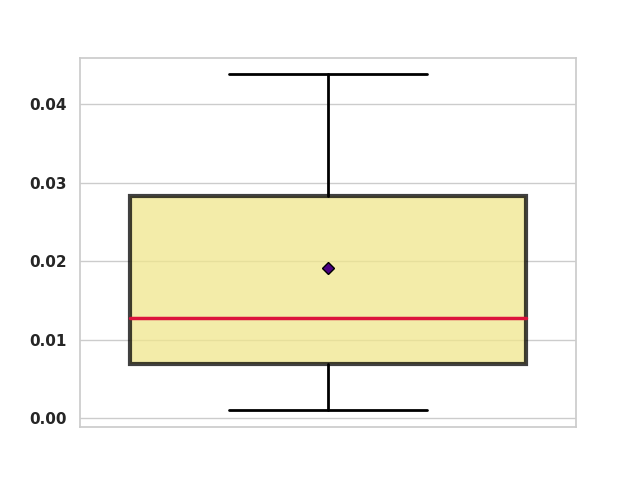}%
\label{fig: rosy 4d_bp}}
\subfloat[]{\includegraphics[width=1.5in]{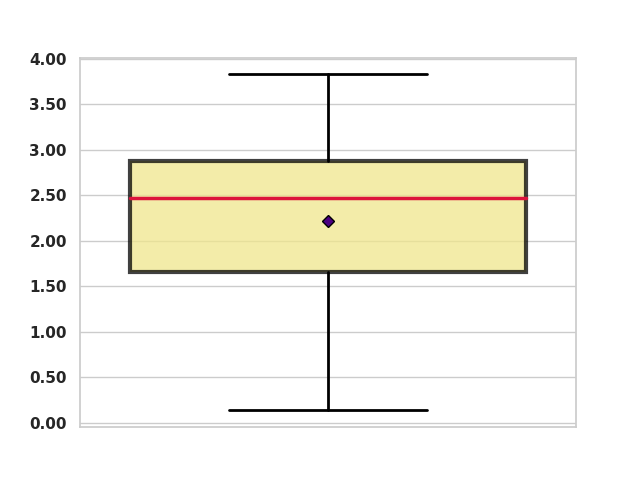}%
\label{fig: schew 4d_bp}}

\subfloat[]{\includegraphics[width=1.5in]{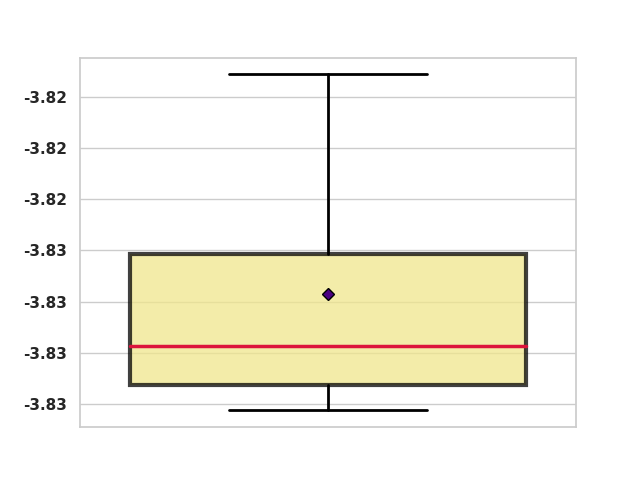}%
\label{fig: vinny 4d_bp}}
\subfloat[]{\includegraphics[width=1.5in]{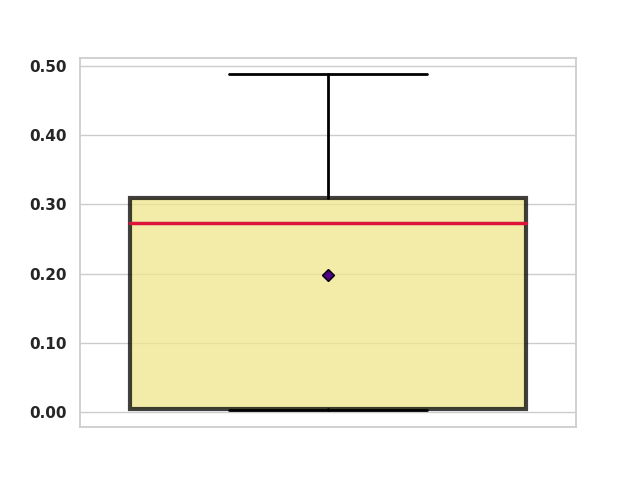}%
\label{fig: xin 4d_bp}}

\caption{Boxplots of fitness distributions over the $30$ independent runs performed for the considered benchmark functions considering $D=4$. The red line represents the median, while the black dot represents the mean. (a) Ackley, (b) Alpine, (c) Griewank, (d) Rastring, (e) Rosenbrock, (f) Schwefel, (g) Vincent, (h) Xin-She Yang n.2.}
\label{fig:bp_dim4}
\end{figure*}

As stated in Section~\ref{subsec: gp-fst-pso}, the goal of this study is to assess the suitability of GP-FST-PSO for optimizing challenging landscapes. 
For each benchmark function described in Section~\ref{subsec: benchmark}, a statistical analysis of the fitness achieved by our method (computed over the $30$ independent runs considered) is reported in Figure~\ref{fig:bp_dim2} (for dimension $D=2$), Figure~\ref{fig:bp_dim3} (for dimension $D = 3$), and Figure~\ref{fig:bp_dim4} (for dimension $D=4$). 
For the sake of clarity, the median value of the fitness is also reported, for all the considered problems' dimensionalities, in Table~\ref{tab:fitness_values}. 
Moreover, for $D=2$, which allows the results to be displayed, some of the benchmark functions are displayed together with the correspondent GP-FST-PSO generated surrogate model (Figure~\ref{fig:fun_approx}).
This visual representation allows the reader to better understand how the proposed method approximates the considered (noisy) landscapes.

Focusing on the Ackley benchmark function, the fitness achieved by GP-FST-PSO is displayed via box-plots for dimension $D=2$ in Figure~\ref{fig: ackley 2d_bp}, for dimension $D=3$ in Figure~\ref{fig: ackley 3d_bp} and, lastly, for dimension $D=4$ in Figure~\ref{fig: ackley 4d_bp}.
The fitness obtained by minimizing the surrogate function (for each of the considered problems' dimensionalities) suggests that the method leads to an excellent approximation of the global optimum of the original function (which is equal to $0$). 
Moreover, experimental results show that the argminimum of the surrogate function corresponds to the minimum of the original fitness landscape, ensuring that the method does not get stuck in local optima.
For dimension $D=2$, it is also possible to observe the approximating function generated by our model: while the shapes of the original Ackley function and the surrogate model are very similar, the latter is devoid of the huge quantity of local optima that characterized Ackley. 

Concerning the Alpine function, experimental results corroborate the suitability of the proposed method. Actually, the global minimum is achieved for all the considered dimensions (Figure~\ref{fig: alpine 2d_bp}, \ref{fig: alpine 3d_bp} and \ref{fig: alpine 4d_bp}), together with the correct argminimum that minimizes the considered function. Furthermore, these results are achieved after a few generations of the search process, thus suggesting that the proposed method can quickly determine a satisfactory approximation of the real function. 

For the Griewank function, conclusions similar to those in the Alpine function case can be drawn. In this situation as well, the proposed method correctly approximates the minimum and argminimum for $D=2$ (Figure~\ref{fig: grie 2d_bp}), $D=3$ (Figure~\ref{fig: grie 3d_bp}), and $D=4$ (Figure~\ref{fig: grie 4d_bp}). Additionally, its convergence to a correct surrogate function is achieved after only a few generations. Figure~\ref{fig: gri_approx} provides an example of a surrogate function. As one can see, despite the similarity in terms of shape between the real and the generated function, the latter does not present the large number of local minima characterizing the Griewank function. 

Regarding the Michalewicz function, results obtained for $D=2$ (Figure~\ref{fig: miche 2d_bp}) show that the method is capable of generating surrogate models that mimic the characteristics of the original function well, as the argminimum turns to be appropriately close to that of the benchmark. However, the surrogate model does not provide a good approximation of the benchmark's minimum. This is because the benchmark, in the neighborhood of the argminimum, is quite steep, implying that small variations in the coordinates lead to a more significant variation of the function's value. 

As for the Rastring function, the method achieves excellent results up to dimension $D=4$, and similar observations as the ones discussed for the Alpine and Griewank functions can be drawn. 

The proposed technique also proves itself successful when dealing with functions in which the minima are not located in $(\vec{0})$, as in the Rosenbrock function. In particular, the method correctly generates a surrogate function able to mimic both the argminimum location and the argminimum value, except for $D=4$, where the method still gets close to the real argminimum location. It is indeed essential to stress that, even if the high fitness value seems to suggest that the method did not converge, this value is just a consequence of the very sharp curvature of the Rosenbrock function around the global optimum, and the obtained minimum is close to the real one. 

For the Schwefel function, the method produces excellent results for $D=2$ and $D=3$, as demonstrated by Figure~\ref{fig: schew 2d_bp} and Figure~\ref{fig: schew 3d_bp} respectively. For $D=4$, the method does not return the exact value of the minimum (which corresponds to $0$). Nevertheless, it correctly locates the argminimum. As for the Rosenbrock function, this result can be explained by taking into account the sharp curvature of the Schwefel function.

Taking into account the Vincent benchmark problem, the results show that GP-FST-PSO can generate a valid surrogate function. In this scenario, the argminimum of the generated model turns out to be relatively quite close to the real location of the argminimum of the original benchmark problem, but, as the function is particularly steep near the argminimum location, the minimum value of the surrogate function differs slightly with respect to the original one. 

Finally, for the Xin-She Yang n.2 function, the method is capable of providing a valid surrogate function for all the considered dimensions. Additionally, as shown in Figure~\ref{fig: xins_approx}, the surrogate model can extrapolate the relevant visual features of the benchmark, such as the location of the different peaks.

\section{Conclusion}
\label{sec: conclusion}

This paper investigates the possibility of solving challenging optimization problems by evolving, via GP, surrogate models that, despite being computationally easier to optimize, preserve the same location for the global optimum.  
To evaluate the fitness of the surrogate models generated during the evolutionary process, we exploited another evolutionary technique, i.e., FST-PSO, to assess the quality of the global minimum of the individual (the aforementioned surrogate function).
The closer the surrogate global minimum gets to the real global minimum of the original noisy benchmark function, the more are the chances for the individual to survive to the next generation. 

The proposed method, called the GP-FST-PSO Surrogate Model, has been tested over different functions, all of them widely used as benchmark problems in the optimization framework.
The method was also tested among different dimensions of the domain space to assert its validity in higher dimensions as well. 
The experimental results show that, in each of the considered benchmark problems, the GP-FST-PSO Surrogate Model achieves a satisfactory outcome.
Overall, GP-FST-PSO results able to generate surrogate models that mimic the shape of the original function and, most importantly, that are characterized by the same location for the global argminimum.
Excellent results are achieved for all the considered dimensions and  benchmarks, thus strengthening the suitability of the proposed method for simplifying the exploration of rugged landscapes.

This work represents the first attempt to create a GP-based surrogate model exploiting the FST-PSO technique to evaluate fitness. 
Considering the promising results obtained in this first analysis, this work paves the way for multiple possible future developments focused on improving the benefits provided by this combination. 
Among the different possibilities, one could rely on a more advanced GP technique for the evolution of the population instead of classic linear GP.
Further, from a more theoretical perspective, one could investigate the behaviour of the GP-FST-PSO Surrogate Model when applied to a less regular function (e.g., continuous or $L^1$) than the ones considered in the present work (all belonging to the $\mathcal{C}^{k}$ class, with $k\geq 1$).

\section*{Acknowledgments}
This work was supported by national funds through the FCT (Funda\c{c}\~ao para a Ci\^encia e a Tecnologia) by the projects GADgET (DSAIPA/DS/0022/2018) and UIDB/04152/2020-Centro de Investiga\c{c}\~ao em Gest\~ao de Informa\c{c}\~ao – MagIC/NOVA IMS. This work was also supported by the financial support from the Slovenian Research Agency (research core funding no. P5-0410).


\begin{thebibliography}{99}

\bibitem{koza1994genetic}
J.~R. Koza, ``Genetic programming as a means for programming computers by
  natural selection,'' \emph{Statistics and computing}, vol.~4, no.~2, pp.
  87--112, 1994.

\bibitem{luke11}
S.~Luke, ``1 essentials of metaheuristicsl,'' 1.

\bibitem{nobile2018fuzzy}
M.~S. Nobile, P.~Cazzaniga, D.~Besozzi, R.~Colombo, G.~Mauri, and G.~Pasi,
  ``Fuzzy self-tuning pso: A settings-free algorithm for global optimization,''
  \emph{Swarm and evolutionary computation}, vol.~39, pp. 70--85, 2018.

\bibitem{cavazzuti2013deterministic}
M.~Cavazzuti, ``Deterministic optimization,'' in \emph{Optimization
  methods}.\hskip 1em plus 0.5em minus 0.4em\relax Springer, 2013, pp. 77--102.

\bibitem{lin2012review}
M.-H. Lin, J.-F. Tsai, and C.-S. Yu, ``A review of deterministic optimization
  methods in engineering and management,'' \emph{Mathematical Problems in
  Engineering}, vol. 2012, 2012.

\bibitem{gomes2019agent}
L.~Gomes, J.~Sp{\'\i}nola, Z.~Vale, and J.~M. Corchado, ``Agent-based
  architecture for demand side management using real-time resources’
  priorities and a deterministic optimization algorithm,'' \emph{Journal of
  Cleaner Production}, vol. 241, p. 118154, 2019.

\bibitem{zhou2018comparative}
H.~Zhou, M.~Song, and W.~Pedrycz, ``A comparative study of improved ga and pso
  in solving multiple traveling salesmen problem,'' \emph{Applied Soft
  Computing}, vol.~64, pp. 564--580, 2018.

\bibitem{sorensen2018history}
K.~S{\"o}rensen, M.~Sevaux, and F.~Glover, ``A history of metaheuristics,'' in
  \emph{Handbook of heuristics}.\hskip 1em plus 0.5em minus 0.4em\relax
  Springer, 2018, pp. 791--808.

\bibitem{papila2002shape}
N.~Papila, W.~Shyy, L.~Griffin, and D.~J. Dorney, ``Shape optimization of
  supersonic turbines using global approximation methods,'' \emph{Journal of
  Propulsion and Power}, vol.~18, no.~3, pp. 509--518, 2002.

\bibitem{feng2006identification}
X.-T. Feng, B.-R. Chen, C.~Yang, H.~Zhou, and X.~Ding, ``Identification of
  visco-elastic models for rocks using genetic programming coupled with the
  modified particle swarm optimization algorithm,'' \emph{International Journal
  of Rock Mechanics and Mining Sciences}, vol.~43, no.~5, pp. 789--801, 2006.

\bibitem{poli2005extending}
R.~Poli, W.~B. Langdon, and O.~Holland, ``Extending particle swarm optimisation
  via genetic programming,'' in \emph{European Conference on Genetic
  Programming}.\hskip 1em plus 0.5em minus 0.4em\relax Springer, 2005, pp.
  291--300.

\bibitem{kanemasa2014algorithm}
M.~Kanemasa and E.~Aiyoshi, ``Algorithm tuners for pso methods and genetic
  programming techniques for learning tuning rules,'' \emph{IEEJ Transactions
  on Electrical and Electronic Engineering}, vol.~9, no.~4, pp. 407--414, 2014.

\bibitem{bhosekar2018advances}
A.~Bhosekar and M.~Ierapetritou, ``Advances in surrogate based modeling,
  feasibility analysis, and optimization: A review,'' \emph{Computers \&
  Chemical Engineering}, vol. 108, pp. 250--267, 2018.

\bibitem{emmerich2002metamodel}
M.~Emmerich, A.~Giotis, M.~{\"O}zdemir, T.~B{\"a}ck, and K.~Giannakoglou,
  ``Metamodel—assisted evolution strategies,'' in \emph{International
  Conference on parallel problem solving from nature}.\hskip 1em plus 0.5em
  minus 0.4em\relax Springer, 2002, pp. 361--370.

\bibitem{jin2005comprehensive}
Y.~Jin, ``A comprehensive survey of fitness approximation in evolutionary
  computation,'' \emph{Soft computing}, vol.~9, no.~1, pp. 3--12, 2005.

\bibitem{ong2005surrogate}
Y.~S. Ong, P.~Nair, A.~Keane, and K.~Wong, ``Surrogate-assisted evolutionary
  optimization frameworks for high-fidelity engineering design problems,'' in
  \emph{Knowledge Incorporation in Evolutionary Computation}.\hskip 1em plus
  0.5em minus 0.4em\relax Springer, 2005, pp. 307--331.

\bibitem{tenne2009model}
Y.~Tenne, ``A model-assisted memetic algorithm for expensive optimization
  problems,'' in \emph{Nature-inspired algorithms for optimisation}.\hskip 1em
  plus 0.5em minus 0.4em\relax Springer, 2009, pp. 133--169.

\bibitem{ulmer2003evolution}
H.~Ulmer, F.~Streichert, and A.~Zell, ``Evolution strategies assisted by
  gaussian processes with improved preselection criterion,'' in \emph{The 2003
  Congress on Evolutionary Computation, 2003. CEC'03.}, vol.~1.\hskip 1em plus
  0.5em minus 0.4em\relax IEEE, 2003, pp. 692--699.

\bibitem{booker1998optimization}
A.~J. Booker, J.~Dennis, P.~D. Frank, D.~B. Serafini, and V.~Torczon,
  ``Optimization using surrogate objectives on a helicopter test example,'' in
  \emph{Computational Methods for Optimal Design and Control}.\hskip 1em plus
  0.5em minus 0.4em\relax Springer, 1998, pp. 49--58.

\bibitem{knill1999response}
D.~L. Knill, A.~A. Giunta, C.~A. Baker, B.~Grossman, W.~H. Mason, R.~T. Haftka,
  and L.~T. Watson, ``Response surface models combining linear and euler
  aerodynamics for supersonic transport design,'' \emph{Journal of Aircraft},
  vol.~36, no.~1, pp. 75--86, 1999.

\bibitem{rai2000improving}
M.~Rai, N.~Madavan, and F.~Huber, ``Improving the unsteady aerodynamic
  performance of transonic turbines using neural networks,'' in \emph{38th
  Aerospace Sciences Meeting and Exhibit}, 2000, p. 169.

\bibitem{madsen2000response}
J.~I. Madsen, W.~Shyy, and R.~T. Haftka, ``Response surface techniques for
  diffuser shape optimization,'' \emph{AIAA journal}, vol.~38, no.~9, pp.
  1512--1518, 2000.

\bibitem{donoho2000high}
D.~L. Donoho \emph{et~al.}, ``High-dimensional data analysis: The curses and
  blessings of dimensionality,'' \emph{AMS math challenges lecture}, vol.~1,
  no. 2000, p.~32, 2000.

\bibitem{giannakoglou2002design}
K.~Giannakoglou, ``Design of optimal aerodynamic shapes using stochastic
  optimization methods and computational intelligence,'' \emph{Progress in
  Aerospace sciences}, vol.~38, no.~1, pp. 43--76, 2002.

\bibitem{ong2003evolutionary}
Y.~S. Ong, P.~B. Nair, and A.~J. Keane, ``Evolutionary optimization of
  computationally expensive problems via surrogate modeling,'' \emph{AIAA
  journal}, vol.~41, no.~4, pp. 687--696, 2003.

\bibitem{manzoni2020surfing}
L.~Manzoni, D.~M. Papetti, P.~Cazzaniga, S.~Spolaor, G.~Mauri, D.~Besozzi, and
  M.~S. Nobile, ``Surfing on fitness landscapes: A boost on optimization by
  fourier surrogate modeling,'' \emph{Entropy}, vol.~22, no.~3, p. 285, 2020.

\bibitem{zhou2006combining}
Z.~Zhou, Y.~S. Ong, P.~B. Nair, A.~J. Keane, and K.~Y. Lum, ``Combining global
  and local surrogate models to accelerate evolutionary optimization,''
  \emph{IEEE Transactions on Systems, Man, and Cybernetics, Part C
  (Applications and Reviews)}, vol.~37, no.~1, pp. 66--76, 2006.

\bibitem{zhou2007memetic}
Z.~Zhou, Y.~S. Ong, M.~H. Lim, and B.~S. Lee, ``Memetic algorithm using
  multi-surrogates for computationally expensive optimization problems,''
  \emph{Soft Computing}, vol.~11, no.~10, pp. 957--971, 2007.

\bibitem{lian2004enhanced}
Y.~Lian, M.-S. Liou, and A.~Oyama, ``An enhanced evolutionary algorithm with a
  surrogate model,'' in \emph{Proceedings of genetic and evolutionary
  computation conference, Seattle, WA}, 2004.

\bibitem{kattan2012evolving}
A.~Kattan and E.~Galvan, ``Evolving radial basis function networks via gp for
  estimating fitness values using surrogate models,'' in \emph{2012 IEEE
  congress on evolutionary computation}.\hskip 1em plus 0.5em minus 0.4em\relax
  IEEE, 2012, pp. 1--7.

\bibitem{holland1992adaptation}
J.~H. Holland, \emph{Adaptation in natural and artificial systems: an
  introductory analysis with applications to biology, control, and artificial
  intelligence}.\hskip 1em plus 0.5em minus 0.4em\relax MIT press, 1992.

\bibitem{kennedy1995particle}
J.~Kennedy and R.~Eberhart, ``Particle swarm optimization,'' in
  \emph{Proceedings of ICNN'95-international conference on neural networks},
  vol.~4.\hskip 1em plus 0.5em minus 0.4em\relax IEEE, 1995, pp. 1942--1948.

\bibitem{zadeh1996fuzzy}
L.~A. Zadeh, ``Fuzzy sets,'' in \emph{Fuzzy sets, fuzzy logic, and fuzzy
  systems: selected papers by Lotfi A Zadeh}.\hskip 1em plus 0.5em minus
  0.4em\relax World Scientific, 1996, pp. 394--432.

\bibitem{takagi1985fuzzy}
T.~Takagi and M.~Sugeno, ``Fuzzy identification of systems and its applications
  to modeling and control,'' \emph{IEEE transactions on systems, man, and
  cybernetics}, no.~1, pp. 116--132, 1985.

\bibitem{code}
G.~Pietropolli, ``Surmod-gp-fst-pso,''
  \url{https://github.com/gpietrop/SurMod-GP-FST-PSO}, 2022.

\bibitem{jamil2013literature}
M.~Jamil and X.-S. Yang, ``A literature survey of benchmark functions for
  global optimisation problems,'' \emph{International Journal of Mathematical
  Modelling and Numerical Optimisation}, vol.~4, no.~2, pp. 150--194, 2013.

\bibitem{liu2019ai}
J.~Liu and J.~Bailey, \emph{AI 2019: Advances in Artificial Intelligence 32nd
  Australasian Joint Conference, Adelaide, SA, Australia, December 2–5, 2019,
  Proceedings}.\hskip 1em plus 0.5em minus 0.4em\relax Springer, 2019.

\bibitem{plevris2022collection}
V.~Plevris and G.~Solorzano, ``A collection of 30 multidimensional functions
  for global optimization benchmarking,'' \emph{Data}, vol.~7, no.~4, p.~46,
  2022.

\bibitem{molga2005test}
M.~Molga and C.~Smutnicki, ``Test functions for optimization needs. test
  functions for optimization needs,'' 2005.

\bibitem{li2013benchmark}
X.~Li, A.~Engelbrecht, and M.~G. Epitropakis, ``Benchmark functions for
  cec’2013 special session and competition on niching methods for multimodal
  function optimization,'' \emph{RMIT University, Evolutionary Computation and
  Machine Learning Group, Australia, Tech. Rep}, 2013.

\bibitem{perkis1994stack}
T.~Perkis, ``Stack-based genetic programming,'' in \emph{Proceedings of the
  First IEEE Conference on Evolutionary Computation. IEEE World Congress on
  Computational Intelligence}.\hskip 1em plus 0.5em minus 0.4em\relax IEEE,
  1994, pp. 148--153.

\end{thebibliography}

\end{document}